# A machine learning platform for development of low flammability polymers

Duy Nhat Phan,*,† Alexander B. Morgan,‡ Lokendra Poudel,† and Rahul Bhowmik*,†

†*Polaron Analytics, 9059 Springboro Pike, Suite C, Miamisburg, 45342, Ohio, USA*
‡*University of Dayton Research Institute, 300 College Park, Dayton, 45469, Ohio, USA*

E-mail: nhat@polaronanalytics.com; rahulbhowmik@polaronanalytics.com

## Abstract

Flammability index (FI) and cone calorimetry outcomes, such as maximum heat release rate, time to ignition, total smoke release, and fire growth rate, are critical factors in evaluating the fire safety of polymers. However, predicting these properties is challenging due to the complexity of material behavior under heat exposure. In this work, we investigate the use of machine learning (ML) techniques to predict these flammability metrics. We generated synthetic polymers using Synthetic Data Vault to augment the experimental dataset. Our comprehensive ML investigation employed both our polymer descriptors and those generated by the RDkit library. Despite the challenges of limited experimental data, our models demonstrate the potential to accurately predict FI and cone calorimetry outcomes, which could be instrumental in designing safer polymers. Additionally, we developed POLYCOMPRED, a module integrated into the cloud-based MatVerse platform, providing an accessible, web-based interface for flammability prediction. This work provides not only the predictive modeling of polymer flammability but also an interactive analysis tool for the discovery and design of new materials with tailored fire-resistant properties.

# Introduction

The prediction of material properties using machine learning (ML) has become increasingly important as researchers seek to accelerate the discovery and optimization of new materials with desired characteristics.[1–3] Within the field of fire safety engineering, understanding the flammability of materials and their behavior under heat exposure is crucial for developing safer products and structures. For instance, understanding how a building material reacts in a fire can make the distinction between containing a fire or enabling it to spread uncontrollably, potentially saving lives and property. While traditional methods of evaluating flammability, such as ASTM E1354 cone calorimetry tests, provide detailed insights into properties like maximum heat release rate, time to ignition, and smoke density, these tests are destructive of the polymer and do require a capital investment that is not all labs can afford. Thus, developing predictive models that can accurately estimate these parameters from material properties is highly desirable.

With advances in machine learning (ML) methods, a growing body of work has leveraged ML techniques to design and predict polymer performance.[4–6] In particular, ML has achieved remarkable successes in generating the prediction of electron affinity (EA) and ionization potential using graph convolutional neural networks (GCNNs).[7] It has also been harnessed to predict shape memory properties through advanced frameworks such as dual-convolutional models[8] and transfer learning-variational autoencoders.[9] In the field of flame retardancy, researchers have applied various ML techniques to predict and optimize the performance of flammable materials. For instance, Nguyen et al.[10] used multiple linear regression (MLR) and Bayesian regularized artificial neural networks with Gaussian prior (BRANNGP) to predict the heat release properties of fiber-reinforced polymer laminates. Bhowmik et al.[11] utilized Decision Tree and Principal Component Analysis methods to predict the specific heat at constant pressure ($C_p$) of polymers. In related work, Chen et al.[12,13] leveraged posynomial modeling along with four other ML approaches—conventional linear regression, nonlinear artificial neural networks, and a combination of Lasso, Ridge, and ANN—to optimize the flame retardancy of polymer nanocomposites. Expanding on these efforts, Chen et al.[14] further applied properties descriptors and regression techniques to design organic phosphorus-containing flame retardant composites. Recently, Yan et al.[15] employed an ML framework based on substructure fingerprinting and self-enforcing deep neural networks (SDNN) to predict the fireproof performance of flame-retardant epoxy resins. Despite these extensive advances, the use of ML to analyze and identify polymer descriptors specifically for predicting the flammability index has yet to be underexplored. This paper addresses this gap by investigating ML approaches to identify key polymer descriptors for predicting flammability index, which is one of the motivations and contributions of this work.

Furthermore, the traditional experiment-based approach to material design, which heavily relies on scientists' domain knowledge, is both costly and time-consuming.[15] While integrating experimental results with simulations of synthetic polymer generation offers a promising approach, its application in polymer science remains unclear. Moreover, most existing ML models for developing low-flammability polymers focus on optimizing the design of flammable material composites but often lack user-friendly interfaces that facilitate end-user interaction. This gap significantly limits the practical application and accessibility of these models, especially for users without technical expertise.

Taken together, this work aims to develop robust ML models to predict the flammability index (FI) and cone calorimetry outcomes, with a focus on maximum heat release rate, time to ignition, total smoke release, and fire growth rate. Our work makes several key contributions to the field of flammability prediction. First, we curated a dataset by extracting ignition temperature and heat of combustion values from the literature for 32 polymers, which we then used to create an FI dataset. We demonstrated how to effectively incorporate expert feedback with ML models to select accurate FI values. Additionally, we conducted fire testing to provide important experimental data and validation for cone calorimetry results. Furthermore, to overcome the limitations of experimental data, which are major challenges in polymer informatics, we generated synthetic polymers using Synthetic Data Vault (SDV), an open-source Python library. To our knowledge, this is the first study to leverage synthetic polymers to enhance polymer datasets.

Besides using our polymer descriptors, we further explored descriptors generated by the RDkit library. Finally, comprehensive analyses were conducted on both real and synthetic datasets to evaluate the proposed models and identify key descriptors for predicting flammability metrics. Importantly, we developed POLYCOMPRED (Polymer Composite Properties Prediction), a module integrated into the cloud-based MatVerse platform that provides powerful tools for data analysis, simulation, and predictive modeling in materials science. By integrating POLYCOMPRED into MatVerse, we provide a user-friendly, web-based interface specifically for flammability prediction. This integration not only makes advanced analysis tools accessible and intuitive for users but also facilitates the discovery of novel material compositions with tailored flammability characteristics, thereby accelerating innovation in polymer design.

In the following sections, we detail our methodology for collecting experimental data and generating synthetic data, along with the development of our proposed ML models for predicting the flammability index and cone calorimetry results. We also describe the components of our interactive web-based interface, including the POLYCOMPRED module. We then present an analysis of the numerical results and conclude with a discussion of the implications of our findings and directions for future work.

# Method

## ML database development

We have focused on our polymer descriptor database, which includes 68 polymers with known experimental specific heat at constant pressure ($C_p$). This database of $C_p$ was established in our previous work.[11] We called our database as Polymer Descriptor Database (PDD). For these polymers, we extracted the ignition temperature $T_i$ and heat of combustion $\Delta H$ from the literature. These values were then used to calculate the flammability index (FI) using the equation provided by Kishore and Mohan Das[16]:

$$FI = C_pT_i/\Delta H \qquad (1)$$

For each polymer, 188 descriptors are extracted from the chemical structure. More details of the descriptors are mentioned in our prior study.[11]

In addition to utilizing our polymer descriptor database, we have also explored descriptors generated by RDKit (https://www.rdkit.org), which is an open-source toolkit for cheminformatics. To better understand the database, polymers were categorized into low, medium, and high flammability index groups. This categorization was conducted in consultation with a fire-retardant materials expert, who selected five polymers for each category. A Random Forest classification model was then trained using these labeled polymers to predict flammability index categories for the remaining polymers. The predicted labels were subsequently used to identify and remove suspect flammability index values.

Due to the challenges in obtaining cone calorimetry data, including maximum Heat Release Rate (pHRR) and Time to Ignition (TIG) values for our polymers, fire testing was conducted on 15 polymers using the ASTM E-1354-23 standard. The Cone Calorimeter experiments were performed at a heat flux of 50 kW/m² with an exhaust flow rate of 24 L/s, in accordance with standardized procedures. However, the samples were not molded specimens, but were instead powders or pellets, so samples were normalized by sample weights rather than sample thicknesses, as is sometimes preferred in cone calorimeter testing. Various indices, including TIG, pHRR, total smoke release, and fire growth rate, were collected (details are provided in the supplemental material).

As applying ML approaches requires a substantial amount of training data to effectively recognize essential features for predictive tasks, we employed the Synthetic Data Vault (SDV)[17] to generate synthetic tabular data. SDV is an open-source Python library that utilizes a variety of machine learning algorithms, ranging from classical statistical methods to Generative AI, to learn patterns and relationships present in real data and emulate them in synthetic data.

## ML model development

We have developed random forest regression models to predict the flammability index, ignition temperature, maximum heat release rate, total smoke release, and fire growth rate. Random forest regression, as a type of ensemble learning, combines the predictions of multiple decision trees to enhance predictive accuracy and mitigate overfitting. This approach is particularly effective in handling the complex, non-linear relationships inherent in material properties. Hyperparameters such as the number of trees (estimators) and tree depth were optimized using cross-validation. The models' performance was evaluated using the $R^2$ score, with a robust assessment conducted through cross-validation. Our ML models were implemented using the Scikit-learn package in Python.[18]

## Model evaluation

All developed models were exclusively trained using synthetic polymers as the training dataset. The performance of these models was subsequently evaluated on both synthetic and real polymers. The accuracy and effectiveness of the models were assessed using the $R^2$ score as the evaluation metric. The $R^2$ score compares the predicted values generated by the models to the actual observed values, providing an indication of how well the models can generalize and predict the properties of testing polymers based on the knowledge acquired from synthetic polymers. An $R^2$ score higher than 0.7 (70%) is considered to indicate good predictive capabilities.[19] The $R^2$ score[20] is defined as follows:

$$R^2 = 1 - \frac{\sum_{i=1}^{n}(y_i - \hat{y}_i)^2}{\sum_{i=1}^{n}(y_i - \bar{y})^2}. \tag{2}$$

Here, $y_1, y_2, ..., y_n$ are the actual (observed) values, $\hat{y}_1, \hat{y}_2, ..., \hat{y}_n$ are the predicted values, and $\bar{y}$ is the mean of the observed values. An $R^2$ score of 1 indicates perfect prediction, where predicted values exactly match observed values. A baseline model that always predicts $\bar{y}$—the average of the observed values—will have an $R^2$ score of 0, as it fails to capture any data variations and simply predicts the mean. Models with predictions worse than this baseline will have a negative $R^2$ score, indicating the predictions are worse than simply using the mean of the observed data.

## Development of an interactive analysis tool

We have developed a module named POLYCOMPRED (Polymer Composite Properties Prediction) within the MatVerse platform (https://matverse.com/) that provides powerful tools for data analysis, simulation, and predictive modeling in materials science. This module includes the descriptor databases and the developed ML methods. Through the webbased interface, users can upload .pdb files and SMILES strings of polymers to predict their flammability metrics. Additionally, users can access visualizations comparing these predicted values with those in our database. Screenshots of the POLYCOMPRED are illustrated in Figures 1, 2, and 3.

Figure 1: Screenshot of POLYCOMPRED.

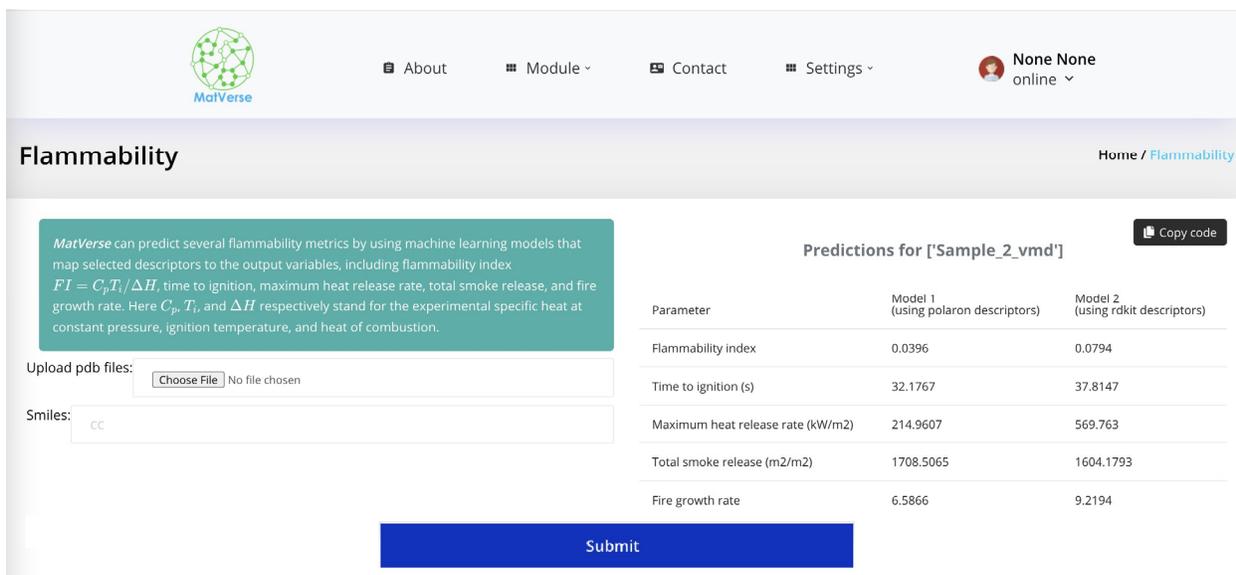

Figure 2: POLYCOMPRED displays predicted flammability results after the submission of a PDB file or SMILES string.

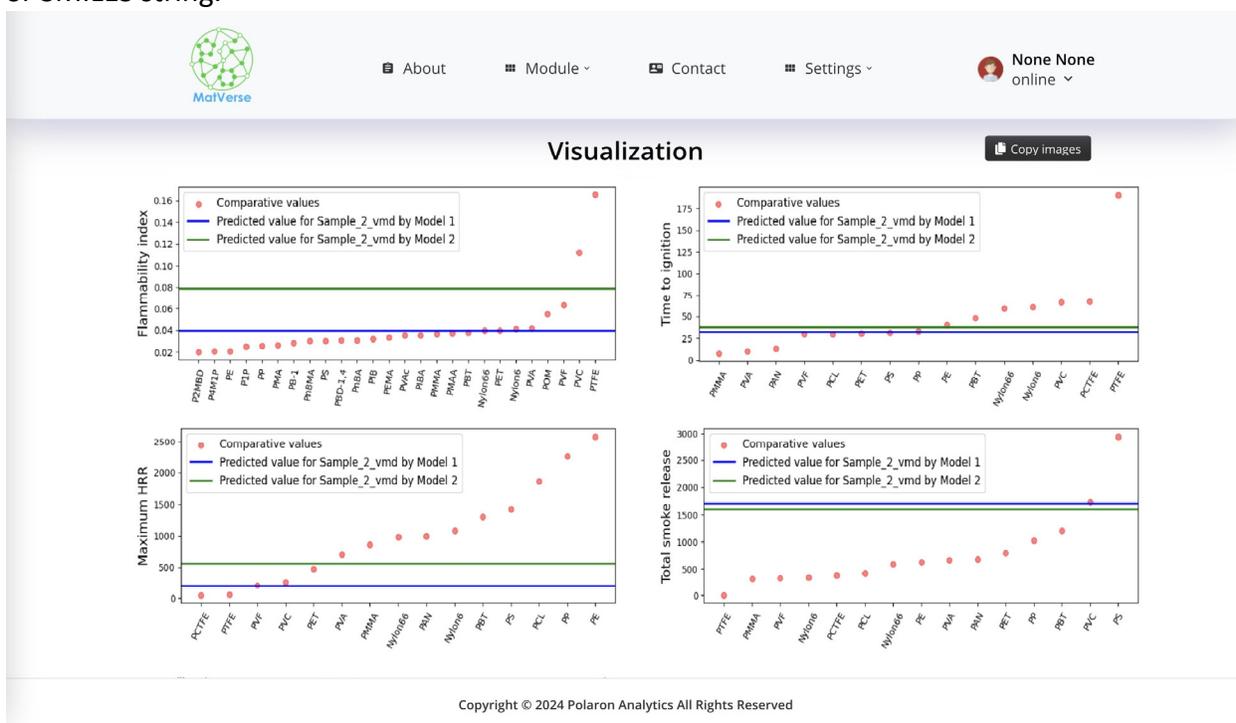

Figure 3: POLYCOMPRED visualizes the predicted flammability results.

# Results and discussion

## Data extraction

We have successfully extracted the ignition temperature $T_i$ and heat of combustion $\Delta H$ values from the literature for 32 out of 68 polymers in our PDD. These values are then used to evaluate their FI values by using the equation (1). Table 1 shows the FI values and predicted labels of 32 polymers. Upon comparing the FI values and labels, we observed that the FI values correspond with their classified labels in most cases. However, three polymers with high FI values have been classified as low, and three polymers with low FI values have been classified as medium or high. Consequently, we removed the six polymers highlighted in yellow from Table 1. The distribution of the remaining 26 polymers is presented in Figure 4.

Table 1: FI values and labels of 32 polymers. We considered High (H) FI values to be within the range of 0.0418 to 0.1652, the Medium (M) FI values within the range of 0.0382 to 0.041, and the Low (L) FI values within the range of 0.0203 to 0.0376. The FI Labels, indicated in green color font in the last column, are used to develop the Random Forest classification, while the red color font represents the predicted labels. The polymer rows highlighted in yellow signify the incorrect FI values, as suggested by the trained ML model.

|   | Name | Mol Wt (g/mol) | $C_p$ (J/mol-K) | $T_i$ (K) | $\Delta H$ (J/g) | FI | FI (label) |
|---|---|---|---|---|---|---|---|
| 1 | Poly(iso-butyl acrylate) | 128.17 | 232.09 | 613.15 | 31380 | 0.0353 | L |
| 2 | Poly(n-butyl acrylate) | 128.17 | 233.28 | 552.03 | 32216.8 | 0.0311 | L |
| 3 | Poly(ethyl acrylate) | 100.12 | 178.88 | 655.92 | 27614.4 | 0.0424 | L |
| 4 | Poly(methyl acrylate) | 86.09 | 151.99 | 733.15 | 23012 | 0.0562 | L |
| 5 | 1,4-Poly(butadiene) | 54.09 | 106 | 693.15 | 44183.04 | 0.0307 | L |
| 6 | Poly(1-butene) | 56.11 | 117.02 | 657.03 | 48460.16 | 0.0282 | L |
| 7 | Poly(ethylene) | 14.03 | 21.81 | 622.05 | 45877.56 | 0.0210 | L |
| 8 | Poly(isobutene) | 56.11 | 110.09 | 738.15 | 44998.92 | 0.0321 | L |
| 9 | Poly(2-methylbutadiene) | 68.12 | 130.2 | 493.15 | 46343.21 | 0.0203 | L |
| 10 | Poly(4-methyl-1-pentene) | 84.16 | 145.4 | 573.15 | 47502.61 | 0.0208 | L |
| 11 | Poly(1-pentene) | 70.14 | 144.34 | 548.15 | 44998.92 | 0.0250 | L |
| 12 | Polypropylene | 42.08 | 68.24 | 736 | 45605.6 | 0.0261 | L |
| 13 | Poly(n-butyl methacrylate) | 142.2 | 263.41 | 567.59 | 34434.32 | 0.0305 | L |
| 14 | Poly(ethyl methacrylate) | 114.15 | 167.42 | 666.48 | 29400 | 0.0332 | L |
| 15 | Poly(methacrylic acid) | 86.09 | 112.5 | 673.15 | 23375.53 | 0.0376 | L |
| 16 | Poly(methacrylamide) | 85.11 | 118.7 | 783.15 | 27345.43 | 0.0399 | L |
| 17 | Poly(methyl methacrylate) | 100.12 | 137.72 | 651 | 24200 | 0.0370 | L |
| 18 | Poly(styrene) | 104.15 | 127.38 | 675 | 27000 | 0.0305 | L |
| 19 | Poly($\alpha$-methylstyrene) | 118.18 | 150.7 | 847.59 | 41128.72 | 0.0262 | L |
| 20 | Poly(acrylonitrile) | 53.06 | 68.83 | 754.15 | 33181.81 | 0.0294 | H |
| 21 | Poly(tetrafluroethylene) | 50.01 | 45.09 | 767 | 4184 | 0.1652 | H |
| 22 | Poly(vinyl chloride) | 62.5 | 59.35 | 675 | 5700 | 0.1124 | H |
| 23 | Poly(vinyl fluoride) | 46.04 | 59.91 | 733.15 | 15062.4 | 0.0633 | H |
| 24 | Poly(vinyl acetate) | 86.09 | 101.86 | 675.372 | 22673.09 | 0.0352 | L |
| 25 | Poly(vinyl alcohol) | 44.05 | 68.11 | 678.15 | 25057.97 | 0.0418 | H |
| 26 | Nylon66 | 226.32 | 331.3 | 785 | 28760 | 0.0399 | M |
| 27 | Nylon6 | 113.16 | 170 | 705.15 | 25800 | 0.0410 | M |
| 28 | Poly(L-methionine) | 131.19 | 176.7 | 486.15 | 24209.16 | 0.0270 | M |
| 29 | Poly(butylene terephthalate) | 220.23 | 355.311 | 633.15 | 26710 | 0.0382 | M |
| 30 | Poly(ethylene terephthalate) | 192.16 | 225.2 | 793.15 | 23220 | 0.0400 | M |
| 31 | Poly(oxymethylene) | 30.03 | 38.52 | 617.15 | 14400 | 0.0549 | H |
| 32 | Poly(4-hydroxybenzoic acid) | 120.11 | 122.6 | 523.15 | 25022.06 | 0.0213 | M |

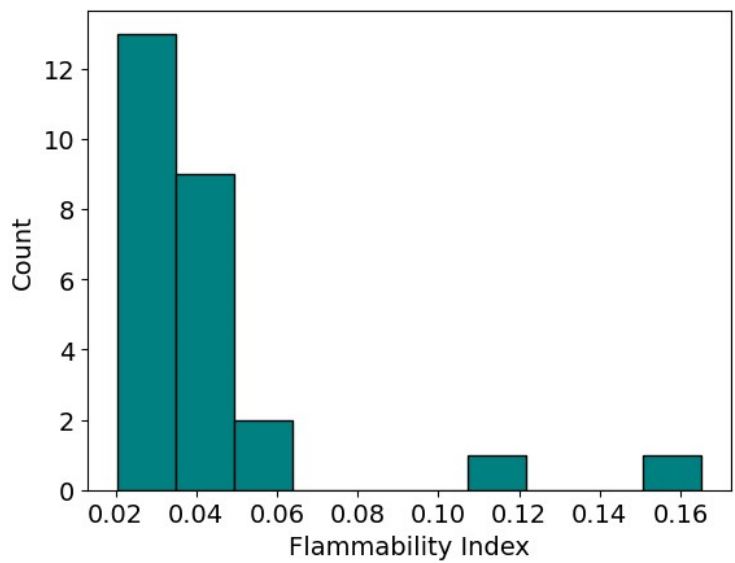

Figure 4: Distribution of FI values of 26 polymers.

Fire testing

We purchased 15 polymers from SIGMA-ALDRICH INC., received in the form of powders, pellets, or film in the case of Polyvinylflouride. We conducted Cone Calorimeter experiments at 1 heat flux (50 kW/m2) with an exhaust flow of 24 L/s using the standardized cone calorimeter procedure (ASTM E-1354-23). We collected various metrics such as time to ignition, maximum heat release rate, total smoke release (TSR), and fire growth rate. The fire growth rate, determined by dividing the peak HRR by the time to peak HRR, represents the rate of fire growth (FIGRA) for a material once exposed to heat. A higher FIGRA suggests faster flame spread and possible ignition of nearby objects. The results are shown in Table 2. The distribution of the 15 polymers is illustrated in Figure 5.

Table 2: The experimental results of 15 polymers.

| Name | TIG | pHRR | TSR | FIGRA |
|---|---|---|---|---|
| Poly(ethylene) | 40.33333 | 2584.667 | 635.3333 | 19.09007 |
| Polypropylene | 33.33333 | 2255.933 | 1028.333 | 17.39718 |
| Poly(methyl methacrylate) | 7.583333 | 866.3667 | 329.6667 | 9.249758 |
| Poly(styrene) | 31.08333 | 1419.733 | 2952 | 14.54145 |
| Poly(acrylonitrile) | 13.5 | 990.3333 | 680.3333 | 13.6418 |
| Poly(chlorotrifluoroethylene) | 67.5 | 48.9 | 389 | 0.337241 |
| Poly(tetrafluroethylene) | 190.6333 | 54.76667 | 4.3 | 0.188348 |
| Poly(vinyl chloride) | 66.66667 | 274.0333 | 1728 | 2.585881 |
| Poly(vinyl fluoride) | 29.66667 | 216.4667 | 339.6667 | 3.216498 |
| Poly(vinyl alcohol) | 10.25 | 704.9 | 669.6667 | 7.498979 |
| Nylon66 | 58.83333 | 986.5667 | 591.6 | 6.638051 |
| Nylon6 | 60.83333 | 1082.7 | 352 | 5.958761 |
| Poly(butylene terephthalate) | 47.66667 | 1302.9 | 1204.333 | 12.83045 |
| Poly(ethylene terephthalate) | 30.41667 | 472.7333 | 801 | 5.95694 |
| Polycaprolactone | 29.75 | 1865.533 | 420.6667 | 19.07601 |

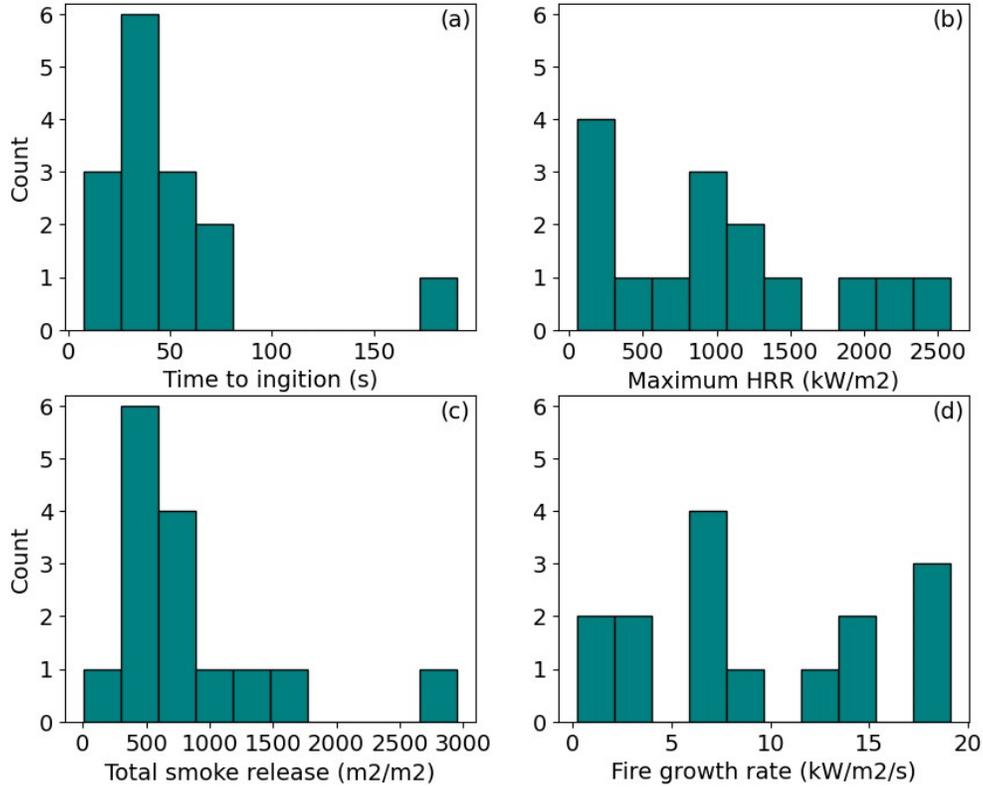

Figure 5: Distribution of TIG (a), pHRR (b), total smoke release (c), and FIGRA (d) values of 15 polymers.

ML model accuracy in predicting flammability

Synthetic Data Generation and Model Performance. To train a machine learning model effectively, sufficient training data is essential for the model to learn the necessary features for predictive tasks. To generate this data, we utilized the SDV. Figure 6 illustrates the distributions of FI, TIG, pHRR, total smoke release, and FIGRA for 1,000 synthetic polymers.

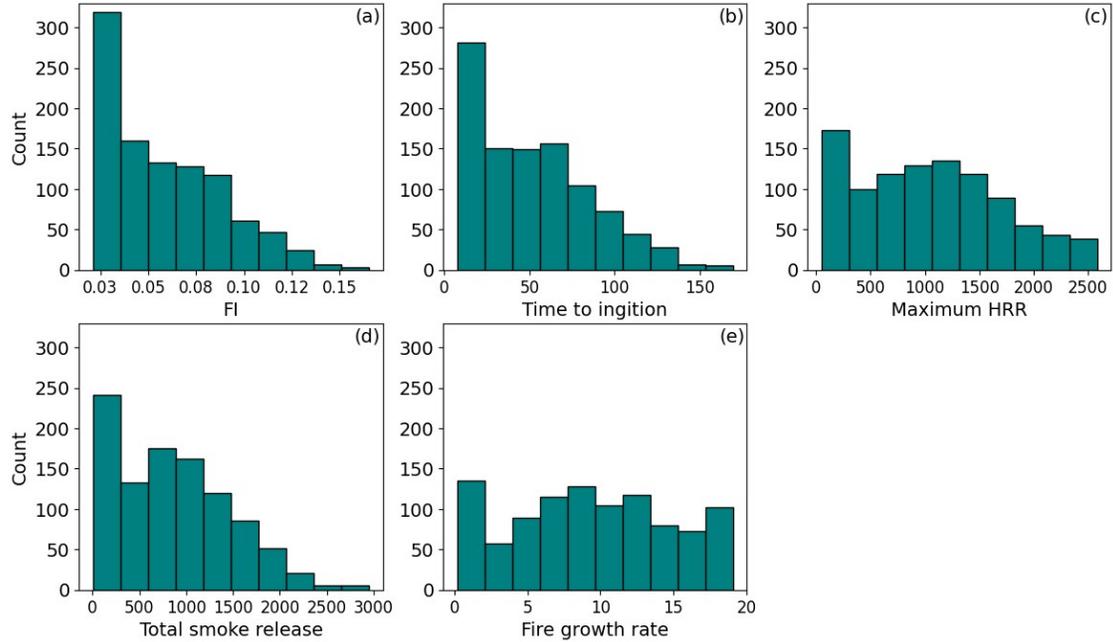

Figure 6: Distribution of FI (a), TIG (b), pHRR (c), total smoke release (d), and FIGRA (e) values of 1,000 synthetic polymers.

To assess the impact of the number of synthetic polymers on model performance, we trained models using varying quantities of synthetic data, ranging from 1,000 to 10,000 polymers. Figure 7 displays the testing $R^2$ scores of the models. Our findings indicate that models trained with data from 7,000, 3,000, 9,000, 5,000, and 6,000 synthetic polymers for predicting FI, TIG, pHRR, total smoke release, and FIGRA, respectively, achieved the highest testing $R^2$ scores of 0.89, 0.84, 0.85, 0.83, and 0.95 on the real polymers.

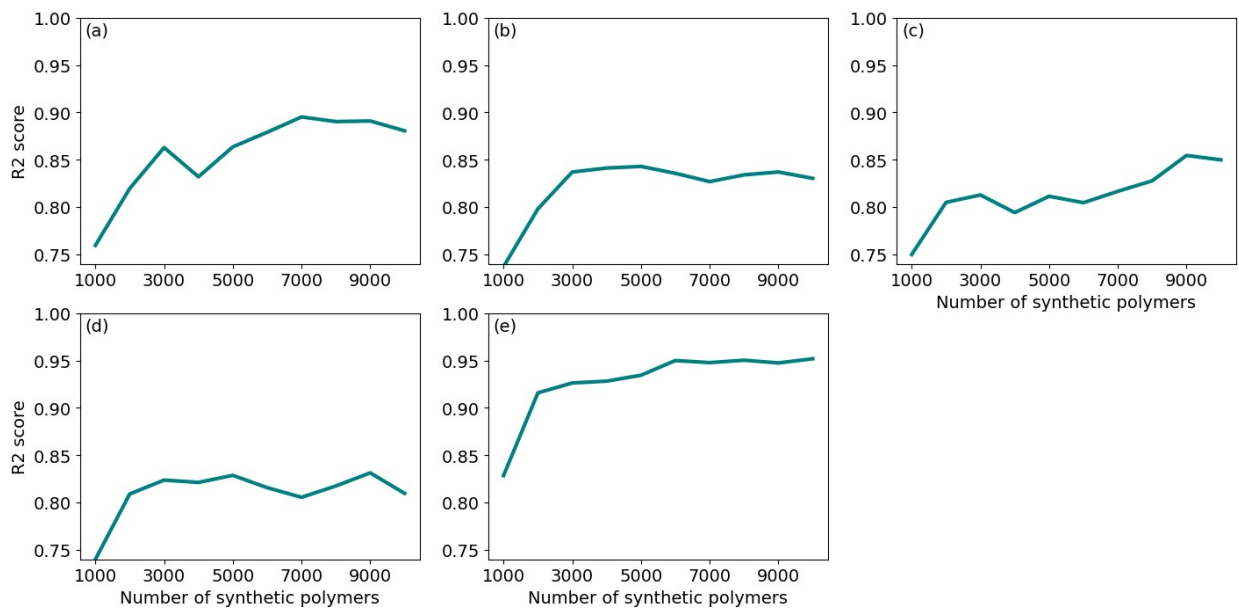

Figure 7: Effect of numbers of synthetic polymers using PDD. Testing $R^2$ scores in predicting FI (a), TIG (b), pHRR (c), total smoke release (d), and FIGRA (e) with respect to the number of synthetic polymers used for training.

Descriptor Selection for Flammability Prediction. We identified significant descriptors that contribute to predicting flammability metrics. To do this, we evaluated the importance of descriptors from models trained using the optimal number of synthetic polymers and 188 descriptors. This process involved assessing the contribution of each feature to reducing impurities—such as Gini impurity or entropy—during data splits. The ten most important descriptors are illustrated in Figure 8 (other important descriptors are provided in the supplemental material).

We then conducted a study to identify the most relevant descriptors for predicting flammability metrics. Various models were trained using different sets of the top descriptors, ranging from 1 to 40. The graph in Figure 9 illustrates the testing $R^2$ scores of these models. In particular, we observed that the model employing the top 20 descriptors achieved the highest testing $R^2$ score of 0.93 for predicting FI values.

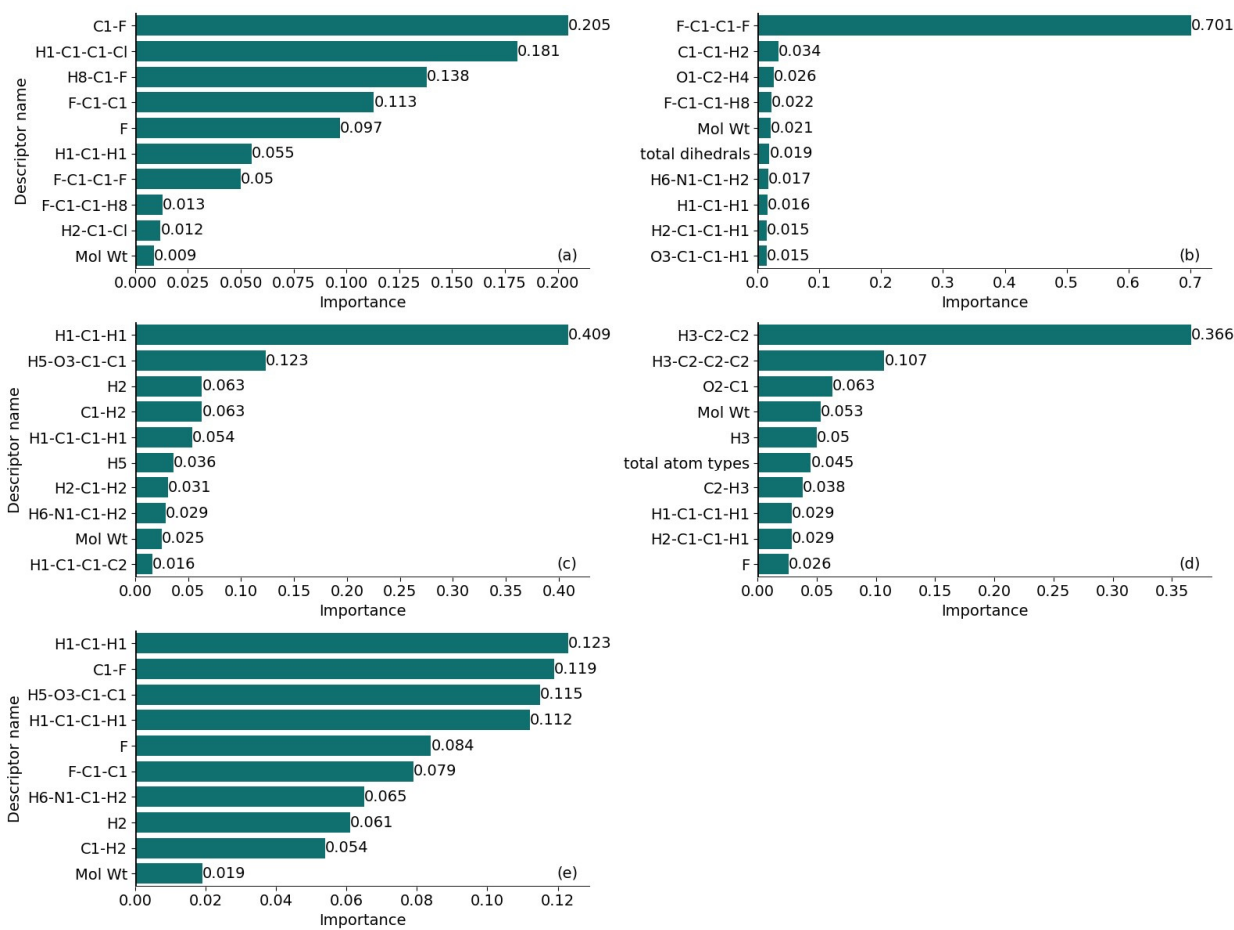

Figure 8: Importance of descriptors in PDD. The top ten important descriptors in predicting FI (a), predicting TIG (b), predicting pHRR (c), predicting total smoke release (d), and predicting FIGRA (e).

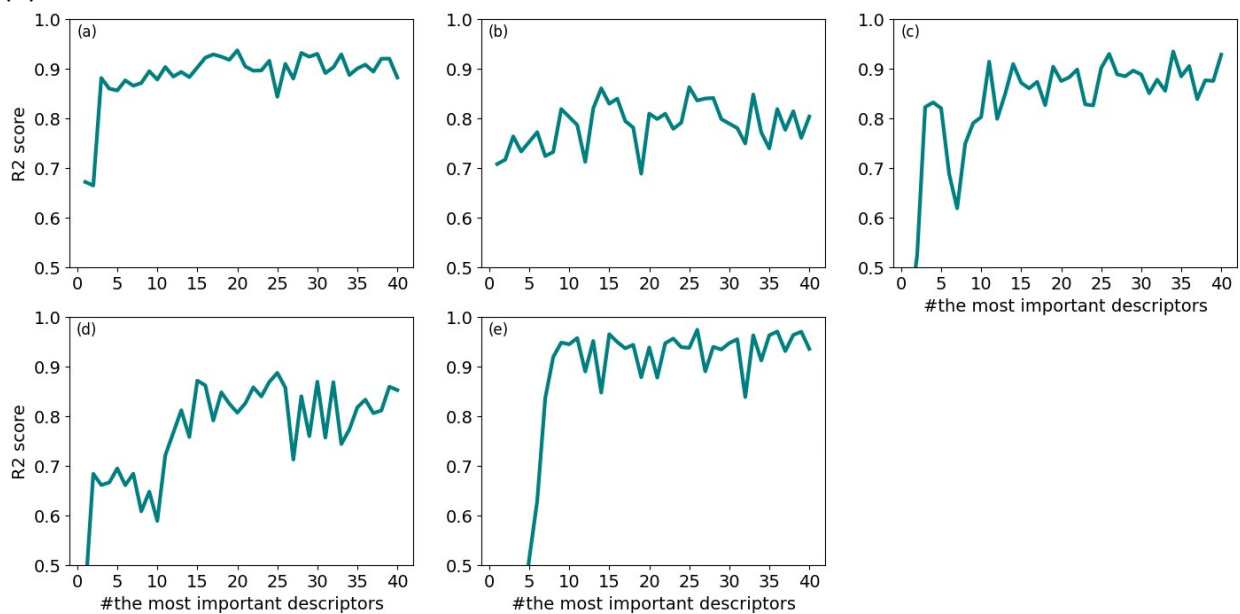

Figure 9: Effect of the polymer descriptors on flammability prediction. Testing $R^2$ scores in predicting FI (a), predicting TIG (b), predicting pHRR (c), predicting total smoke release (d), and predicting FIGRA (e) with respect to the number of the most important descriptors.

Table 3 displays the training and testing of our final ML models in predicting flammability metrics.

Table 3: Training and testing results of the final ML models.

| Metric | Training $R^2$-score on synthetic data | Testing $R^2$-score on real data |
| --- | --- | --- |
| FI | 0.9879 | 0.9365 |
| Time to ignition | 0.9760 | 0.8624 |
| Maximum HRR | 0.9808 | 0.9342 |
| Total smoke release | 0.9666 | 0.8867 |
| Fire growth rate | 0.9829 | 0.9736 |

Based on the results in Table 3, our random forest regression models using PDD demonstrate strong predictive capabilities across flammability metrics. The training $R^2$ scores indicate an excellent model fit, with FI (0.98), time to ignition (0.97), maximum HRR (0.98), total smoke release (0.96), and fire growth rate (0.98) all showing high accuracy. When evaluated on the testing set, the models maintained substantial predictive power, with FI and maximum HRR both achieving an $R^2$ score of 0.93, total smoke release at 0.88, and fire growth rate at 0.97. Although the model's performance for time to ignition decreased to 0.86 in the testing phase, it still reflects a reliable level of accuracy. These results suggest that while the polymer descriptors provide robust model training, they also generalize well to new polymers, particularly for metrics like maximum HRR and fire growth rate, making them effective for predicting flammability characteristics.

## Exploring RDKit descriptor database

To explore other descriptors and further improve our models, we utilized RDKit, an opensource cheminformatics toolkit. This library can generate 210 descriptors, including the number of rotatable bonds, heavy atoms, hydrogen bond acceptors, hydrogen bond donors, molecular weight, and more. For further details, refer to https://www.rdkit.org. Some of the descriptors generated by the RDKit toolkit for 20 polymers are illustrated in Figure 1 of the supplemental material.

We applied the same procedure used with PDD as described above. Specifically, we determined the optimal number of synthetic polymers for training each model, as shown in Figure 10. We also identified the most important descriptors for predicting flammability metrics, as illustrated in Figures 11 and 12. Figure 11 shows the top ten important RDKit descriptors (other important descriptors are provided in the supplemental material) while Figure 12 provides testing $R^2$ scores in predicting flammability metrics with respect to the number of the most important RDKit descriptors.

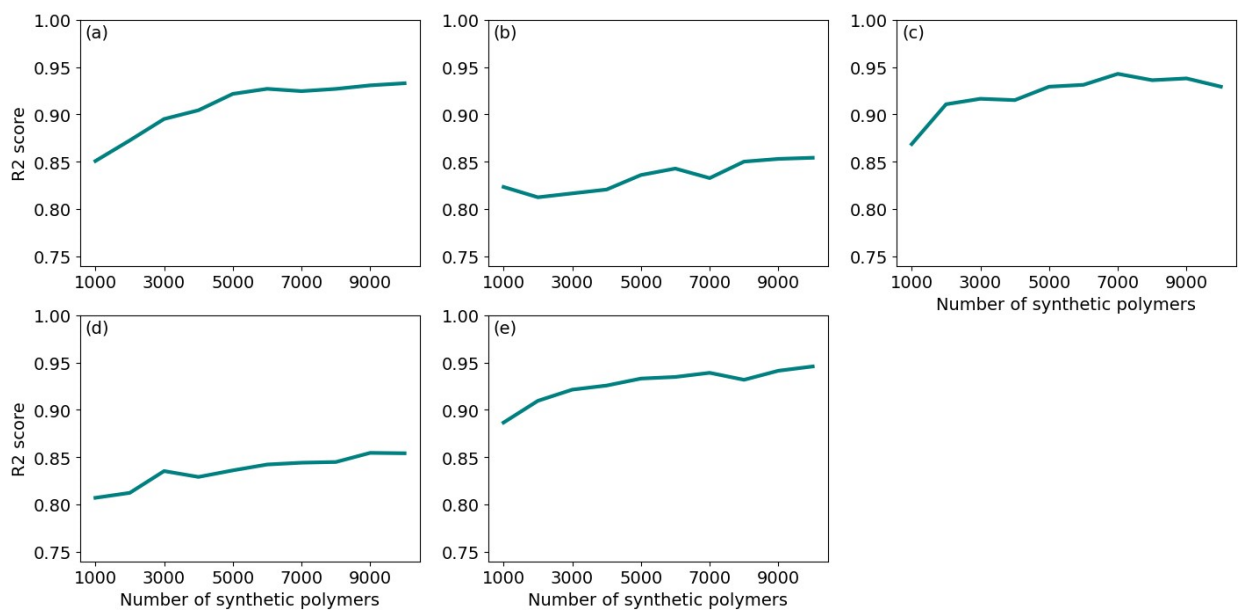

Figure 10: Effect of numbers of synthetic polymers using RDKit descriptors. Testing $R^2$ scores in predicting FI (a), predicting TIG (b), predicting pHRR (c), predicting total smoke release (d), and predicting FIGRA (e) with respect to the number of synthetic polymers used for training.

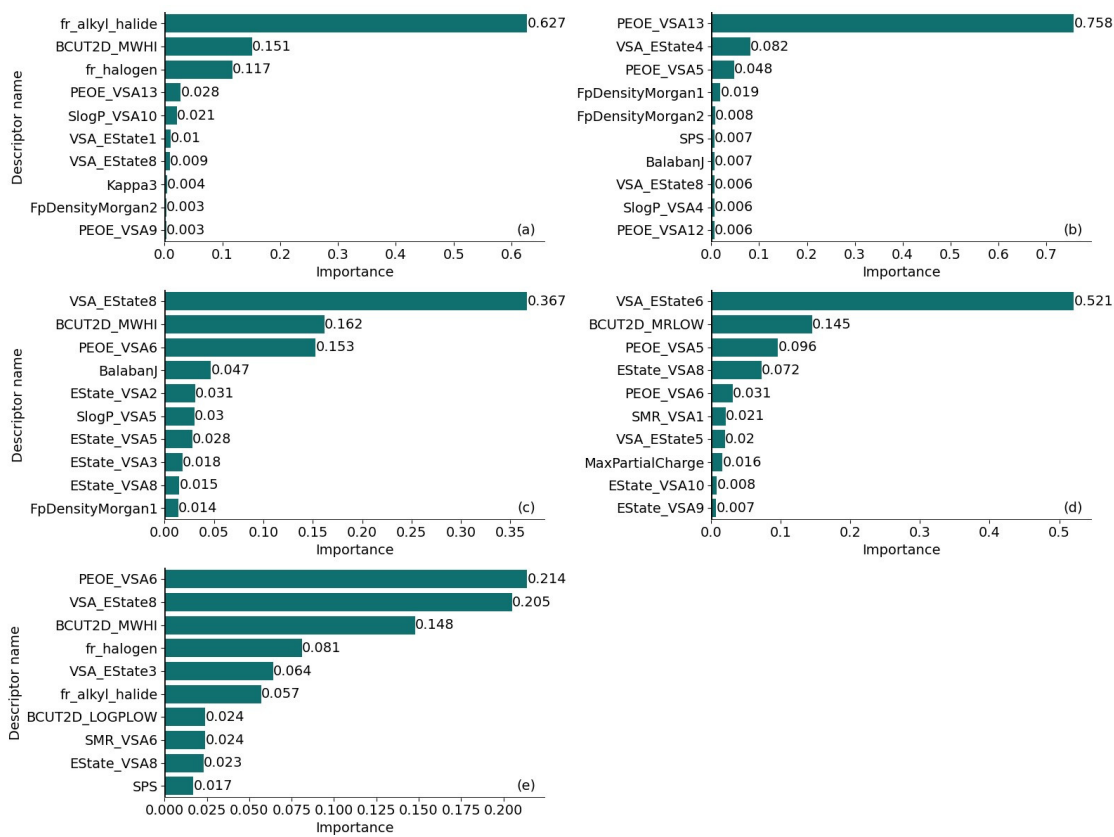

Figure 11: RDKit descriptor importance. The top ten important descriptors in predicting FI (a), predicting TIG (b), predicting pHRR (c), predicting total smoke release (d), and predicting FIGRA (e).

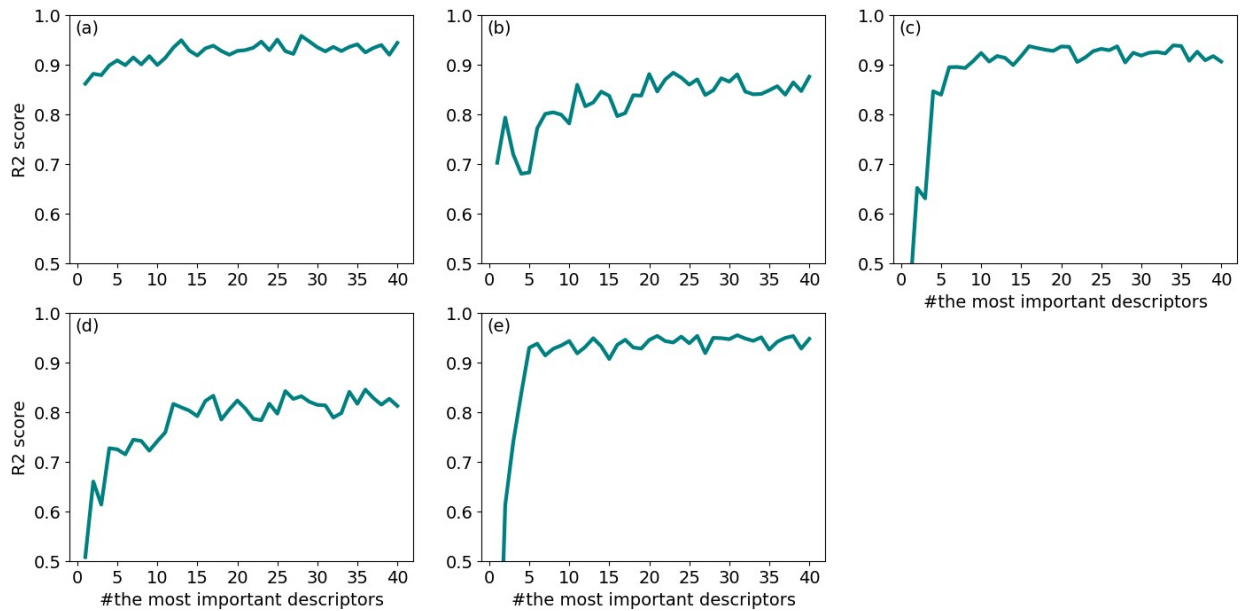

Figure 12: Effect of RDKit descriptors on flammability prediction. Testing $R^2$ scores in predicting FI (a), predicting TIG (b), predicting pHRR (c), predicting total smoke release (d), and predicting FIGRA (e) with respect to the number of the most important RDKit descriptors.

Table 4 summarizes the performance of our random forest regression models using two different sets of descriptors — polymer descriptors and RDKit descriptors—to predict flammability metrics. The results are reported as $R^2$ scores for both training and testing datasets.

Table 4: The results of our random forest regression models with polymer and RDKit descriptors.

| Metric | Training $R^2$-score on synthetic data | | Testing $R^2$-score on real data | |
| --- | --- | --- | --- | --- |
| | Polymer descriptors | RDKit descriptors | Polymer descriptors | RDKit descriptors |
| FI | 0.9879 | 0.9951 | 0.9365 | 0.9576 |
| Time to ignition | 0.9760 | 0.9917 | 0.8624 | 0.8833 |
| Maximum HRR | 0.9808 | 0.9848 | 0.9342 | 0.9157 |
| Total smoke release | 0.9666 | 0.9926 | 0.8867 | 0.8451 |
| Fire growth rate | 0.9829 | 0.9881 | 0.9736 | 0.9546 |

In our analysis, we evaluated the performance of random forest regression models using two sets of descriptors, RDKit and polymer, across various flammability metrics. During the training phase, models utilizing RDKit descriptors consistently outperformed those using polymer descriptors, as indicated by higher $R^2$ scores for Flammability Index, Time to Ignition, Maximum Heat Release Rate, Total Smoke Release, and Fire Growth Rate. However, in the testing phase, models with RDKit descriptors provided more accurate predictions for FI and Time to Ignition, while those with polymer descriptors excelled in predicting Maximum HRR, Total Smoke Release, and Fire Growth Rate. Importantly, the testing $R^2$ scores suggest that models using polymer descriptors may be more robust for certain metrics, potentially reducing the risk of overfitting on real polymers compared to those using RDKit descriptors.

## Cross Validation

To avoid overfitting and provide a more realistic estimate of the model's generalization performance, we generated a synthetic training set and a synthetic testing set for each model. We trained the models on the training tests and evaluated them on both synthetic testing sets and real testing sets. We repeated the process 10 times and reported the average results in Table 5.

Table 5: The results of our random forest regression models with Polymer and RDKit descriptors.

| Metric | Average $R^2$-score on real testing data | | Average $R^2$-score on synthetic testing data | |
|---|---|---|---|---|
| | Polymer descriptors | RDKit descriptors | Polymer descriptors | RDKit descriptors |
| FI | 0.9337 | 0.9529 | 0.9266 | 0.9812 |
| Time to ignition | 0.8652 | 0.8821 | 0.8520 | 0.9513 |
| Maximum HRR | 0.9375 | 0.9157 | 0.9490 | 0.9042 |
| Total smoke release | 0.8939 | 0.8473 | 0.8891 | 0.9531 |
| Fire growth rate | 0.9724 | 0.9582 | 0.9347 | 0.9213 |

Refer to Table 5, where our machine learning models utilizing Polymer descriptors or RDKit descriptors achieved impressive accuracies. In fact, the models using RDKit descriptors even yielded better results on synthetic testing data than on real testing data in terms of FI, time to ignition, and total smoke release.

## Discussions and Conclusion

In this work, we developed random forest regression models utilizing PDD to predict various flammability metrics, including flammability index (FI), maximum heat release rate, time to ignition, total smoke release, and fire growth rate. We created an FI dataset by extracting ignition temperature and heat of combustion values for 32 polymers from the literature. To overcome the limitations of experimental data, we introduced the use of SDV to generate synthetic polymers, enhancing the dataset and improving model generalization. This approach demonstrates the potential of synthetic data to advance predictive modeling in applied materials. We also explored additional descriptors generated by the RDkit library, broadening our analysis. Comprehensive analyses on both real and synthetic datasets allowed us to evaluate model performance and identify key descriptors for accurately predicting flammability metrics.

Importantly, we developed POLYCOMPRED (Polymer Composite Properties Prediction), a module integrated into the MatVerse platform. This module provides a tool for data analysis, simulation, and predictive modeling, specifically tailored for predicting flammability in polymers. POLYCOMPRED features an interactive web-based interface that allows users to input chemical structures and receive accurate predictions of key flammability parameters, including FI value, time to ignition, maximum heat release rate, total smoke release, and fire growth rate. This integration enhances the usability and accessibility of flammability prediction tools, making them available for practical applications in polymer science. Moreover, it enables users to discover novel material compositions with tailored flammability characteristics, which accelerates innovation in polymer design.

Our results demonstrated robust predictive capabilities, with high $R^2$ scores in both training and testing phases, highlighting the effectiveness of polymer descriptors in predicting flammability characteristics. Moreover, expanding the dataset with synthetic data from SDV significantly enhanced model generalization and accuracy. These findings make our models a promising approach for developing safer materials.

Our work has significant implications for applied material design, particularly in enhancing safety and performance. Integrating our models into the POLYCOMPRED module streamlines and facilitates the flammability prediction process, making it accessible to researchers without programming expertise. This also enables the efficient development of materials with tailored fire-resistant properties, essential for advancing safety standards in industries such as construction, automotive, and consumer goods. POLYCOMPRED provides an accessible resource, which allows researchers and industry professionals to innovate and improve material safety, ultimately leading to more resilient and secure products.

Our study highlights the need to further enhance model generalization by expanding the descriptor set and incorporating additional data sources. Future work will include extending the range of polymers tested, applying our models to real-world scenarios, and testing the usability of the POLYCOMPRED tool. Additionally, we will explore using predicted cone calorimeter results to estimate other key measurements of material fire hazards, such as fire growth potential, product fire hazard, and material fire hazard,[21] or training our models to directly predict these values. These efforts will enhance the reliability and applicability of our models while ensuring the tool is user-friendly and effective for researchers and industry professionals. By continuing to refine our approach and assess the tool's usability, we aim to maintain adaptability and robustness, contributing to ongoing advancements in material safety and performance.

# Acknowledgement

This material is based upon work supported by the Office of Naval Research under Contract No. N68335-24-C-0124.